\documentclass{article}

\usepackage{arxiv}

\usepackage[utf8]{inputenc} 
\usepackage[T1]{fontenc}
\usepackage{hyperref}
\usepackage{url}
\usepackage{booktabs}
\usepackage{amsfonts}
\usepackage{nicefrac}
\usepackage{microtype}
\usepackage{graphicx}
\usepackage{natbib}
\usepackage{doi}
\usepackage{multirow}
\usepackage{caption}
\usepackage{subcaption}
\usepackage{amsmath}

\title{When Answers Stray from Questions: Hallucination Detection via Question-Answer Orthogonal Decomposition}

 \author{ {Siyang Yao}\\
 	Shanghai Jiao Tong University\\
 	Shanghai,China\\
 	\And
	{Erhu Feng}\\
 	Shanghai Jiao Tong University\\
 	Shanghai,China\\
    \And
	{Yubin Xia}\\
 	Shanghai Jiao Tong University\\
 	Shanghai,China\\
}

\date{}

\renewcommand{\headeright}{}
\renewcommand{\undertitle}{}

\begin{document}
\maketitle

\begin{abstract}
Hallucination detection in large language models (LLMs) requires balancing accuracy, efficiency, and robustness to distribution shift. Black-box consistency methods are effective but demand repeated inference; single-pass white-box probes are efficient yet treat answer representations in isolation, often degrading sharply under domain shift. We propose \textsc{QAoD} (\textbf{Q}uestion-\textbf{A}nswer \textbf{O}rthogonal \textbf{D}ecomposition), a single-pass framework that projects away the question-aligned direction from the answer representation to obtain a question-orthogonal component that suppresses domain-conditioned variation. To identify informative signals, \textsc{QAoD} further selects layers via diversity-penalized Fisher scoring and discriminative neurons via Fisher importance. To address both in-domain detection and cross-domain generalization, we design two complementary probing strategies: pairing the orthogonal component with question context yields a joint probe that maximizes in-domain discriminability, while using the orthogonal component alone preserves domain-agnostic factuality signals for robust transfer. \textsc{QAoD}'s joint probe achieves the best in-domain AUROC across all evaluated model-dataset pairs, while the orthogonal-only probe delivers the strongest OOD transfer, surpassing the best white-box baseline by up to 21\% on BioASQ at under 2\% of generation cost.
\end{abstract}

\section{Introduction}

Despite their remarkable progress in natural language generation and reasoning, large language
models (LLMs) remain vulnerable to hallucination, the production of factually unsupported or
fabricated content \citep{Huang_2025,ji2023survey,tonmoy2024comprehensive,zhang2025sirenssongaiocean}.
This limitation continues to hinder the safe use of LLMs in high-stakes settings such as healthcare
and legal services \citep{wang2024factualitylargelanguagemodels,Azamfirei2023}.

A common strategy for hallucination detection is to sample multiple candidate answers and compare
their consistency. Although effective, such black-box methods typically require repeated inference
and therefore incur substantial computational overhead \citep{sindex,selfcheckgpt,semanticentropy}.
This efficiency bottleneck has motivated a second line of work that aims to detect hallucinations
from a single generation by probing internal states of the model \citep{SAPLMA,MIND,LLM-Check}.
These single-pass white-box methods are considerably faster, but many of them rely primarily on
answer-side representations and may be sensitive to domain shift or to how factuality-related
signals are entangled with contextual priors \citep{Levinstein_2024}.

We take a third path: rather than comparing across multiple sampled 
outputs or probing the answer representation alone, \textsc{QAoD} 
(\textbf{Q}uestion-\textbf{A}nswer \textbf{O}rthogonal \textbf{D}ecomposition) 
exploits the geometric relationship between question and answer 
representations within a single forward pass. Through this geometric 
decomposition, we realize the following contributions:

\begin{itemize}
\item \textbf{Geometric decoupling for robust representations}: We introduce $\mathcal{H}_{V_{\perp}}$ by removing the question-aligned component from answer representations, suppressing domain variation while preserving the deviation signal. A joint variant $\mathcal{H}_{Q \oplus V_{\perp}}$ additionally incorporates question context for improved in-domain performance.

\item \textbf{Fisher-based layer and neuron selection}: We propose a Fisher-discriminant criterion to identify informative layers and neurons in a single pass over training features, retaining the most discriminative signals without iterative optimization.

\item \textbf{Efficient detection for in-domain and cross-domain settings}: The joint probe achieves the best in-domain AUROC on all evaluated model-dataset pairs and the orthogonal-only probe surpasses the strongest white-box baseline by up to 21\% AUROC points on zero-shot BioASQ. Both operate in a single forward pass at under 2\% of generation cost.
\end{itemize}

\section{Related work}

\subsection{Black-box methods: uncertainty estimation}

Hallucination detection has been studied extensively through output probabilities, repeated sampling, and semantic consistency. SelfCheckGPT \citep{selfcheckgpt} detects hallucinations by sampling multiple responses and measuring their mutual consistency. Semantic entropy \citep{semanticentropy} addresses the limitation that surface-level token diversity can obscure meaning-level agreement: it clusters sampled responses by bidirectional entailment and computes entropy over semantic classes, while SINdex \citep{sindex} extends this with cosine-similarity-based clustering and coherence-weighted entropy. Semantic Dispersion \citep{lin2024generatingconfidenceuncertaintyquantification}, Self-Prompt \citep{languagemodelsmostlyknow}, direct verbalization \citep{lin2022teaching}, and ranked-voting self-consistency \citep{wang-etal-2025-ranked} offer further variants. These approaches are effective, but require multiple inference passes, incurring substantial computational overhead \citep{selfcheckgpt,chern2023factoolfactualitydetectiongenerative,languagemodelsmostlyknow}.

\subsection{White-box methods: internal state-based probing}

White-box methods instead probe hidden states for factuality-related signals \citep{belinkov-2022-probing}. Prior work has shown that factual knowledge can appear as a linear direction in hidden space \citep{implicitrepresentationsmeaningneural,geometrytruthemergentlinear,zou2025representationengineeringtopdownapproach}, and that such internal signals can be exploited for probing even when the model output is hallucinated \citep{li2024inferencetimeinterventionelicitingtruthful}. Existing methods include lightweight classifiers on hidden states \citep{SAPLMA,MIND}, layer-comparison strategies such as DoLa \citep{DoLa}, and other layer-wise or contrastive probes that aggregate intermediate representations across layers \citep{kim2025detectingllmhallucinationlayerwise,LLM-Check,chen2024hallucinationdetectionrobustlydiscerning,chen2024insidellmsinternalstates}. Despite these advances, white-box methods can still entangle domain-sensitive contextual signals with factuality-related signals, which may weaken OOD generalization \citep{Levinstein_2024}. A concurrent work, MHAD \citep{MHAD}, also pools question and answer representations across multiple types and timesteps, but combines them via direct concatenation; this mechanical aggregation does not separate factuality-relevant deviation from question-conditioned context, which reduces its robustness under domain shift. As MHAD's source code is unreleased and several implementation details are underspecified, we could only reproduce it from the paper description and therefore do not include it as a formal baseline; our reproduction trails \textsc{QAoD} on all four evaluated LLMs on zero-shot BioASQ, with an average gap of approximately 10\% AUROC points. \textsc{QAoD} improves OOD robustness by explicitly projecting out the question-aligned component from the answer representation and probing the remaining question-orthogonal component.

\section{Methodology}

\begin{figure}[!ht]
    \centering
    \includegraphics[width=0.9\linewidth]{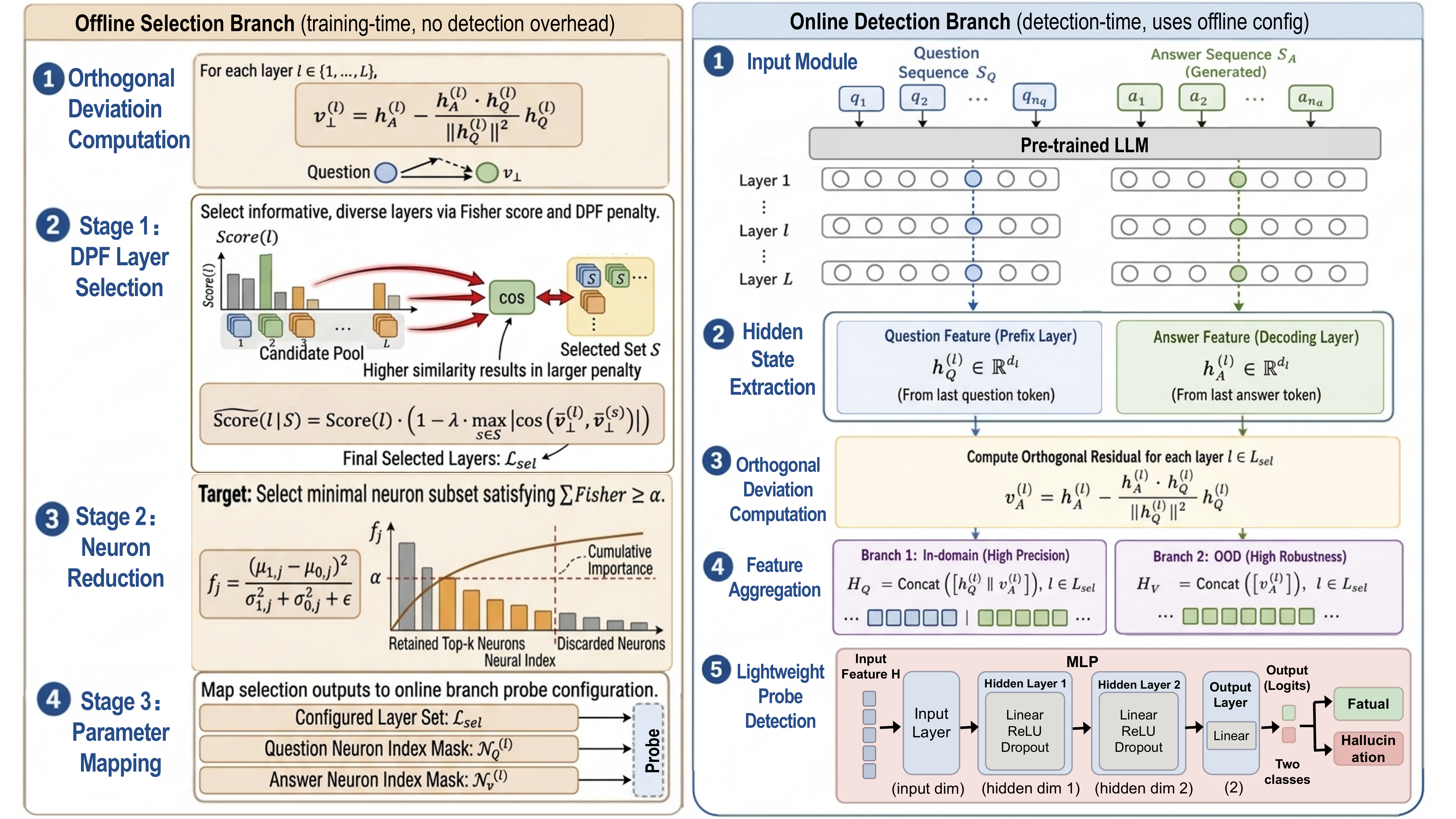}
    \caption{\textsc{QAoD} architecture: an offline selection branch identifies discriminative layers and neurons via Fisher-based scores; the online backbone extracts selected features and computes question-orthogonal components in a single forward pass.}
    \label{fig:architecture}
\end{figure}

The hallucination detection framework \textsc{QAoD} consists of two branches. The offline branch identifies informative layers and neurons during training, while the online branch performs single-pass detection at test time. Figure~\ref{fig:architecture} provides an overview of this offline-online pipeline.

\subsection{Hidden state extraction}

For question sequence $S_Q$ and answer sequence $S_A$, we extract the final-token hidden states at each transformer layer $l$, denoted $h_Q^{(l)}, h_A^{(l)} \in \mathbb{R}^d$. In decoder-only models, causal self-attention ensures that the final token aggregates information from all preceding tokens \citep{vaswani2023attentionneed}, making it a natural read-out point; this choice is further supported by prior probing work showing its effectiveness for factuality detection \citep{SAPLMA,MIND,li2024inferencetimeinterventionelicitingtruthful}.

\subsection{Orthogonal deviation computation}

For each transformer layer $l \in \{1, \dots, L\}$, we compute the question-orthogonal component $v_{\perp}^{(l)}$ by removing from the answer representation its component aligned with the question representation:
\begin{equation}
\label{eq:vperp}
v_{\perp}^{(l)} = h_A^{(l)} - \mathrm{proj}_{h_Q^{(l)}}(h_A^{(l)}) = h_A^{(l)} -
\frac{h_A^{(l)} \cdot h_Q^{(l)}}{\|h_Q^{(l)}\|^2} h_Q^{(l)}.
\end{equation}
The projection is a sample-wise decomposition step: it removes the component of $h_A^{(l)}$ collinear with the current question direction, leaving a question-orthogonal component expected to attenuate question-conditioned domain variation. Section~\ref{sec:ood} empirically examines whether this orthogonal decomposition improves OOD stability.

\subsection{Fisher-discriminant layer and neuron selection}
\label{sec:fisher_selection}

Given the significant depth of contemporary LLMs, naively concatenating hidden state vectors from all layers introduces severe parameter redundancy and overfitting risk. We therefore adopt a Fisher discriminant scoring procedure that operates on class-conditional statistics to identify the most informative layers and neurons for detection.

\subsubsection{Layer scoring via multivariate Fisher discriminant}

For each candidate layer $l$, we compute Fisher scores over the question representations $h_Q^{(l)}$ and the question-orthogonal components $v_\perp^{(l)}$. We use the score
\begin{equation}
    F(X,y)=\frac{\|\mu_1-\mu_0\|^2}{\mathrm{tr}(\Sigma_1)+\mathrm{tr}(\Sigma_0)}
\end{equation}
where $\mu_c$ and $\Sigma_c$ are the class-conditional mean and covariance estimated per label group from the training set. In our implementation, the denominator uses a per-dimension diagonal variance approximation. The layer score is $\text{Score}(l) = F\!\left(v_\perp^{(l)},\, y\right)$ for $\mathcal{H}_{V_\perp}$, and $\text{Score}(l) = \bigl(F\!\left(h_Q^{(l)},\, y\right) + F\!\left(v_\perp^{(l)},\, y\right)\bigr)/2$ for $\mathcal{H}_{Q \oplus V_\perp}$.

\paragraph{Diversity-penalized Fisher (DPF) layer selection.}
Pure top-$K$ Fisher ranking tends to select consecutive layers; since adjacent layers share high inter-layer representational similarity, the resulting selection encodes largely redundant information while missing complementary signals from distinct processing stages. We therefore employ a greedy \emph{Diversity-Penalized Fisher} (DPF) procedure: given the set $S$ of already-selected layers, each candidate $l$ is scored as
\begin{equation}
    \label{eq:dpf}
    \widetilde{\text{Score}}(l \mid S) = \text{Score}(l) \cdot
    \left(1 - \lambda \cdot \max_{s \in S}
    \left|\cos\!\left(\bar{v}_\perp^{(l)},\, \bar{v}_\perp^{(s)}\right)\right|\right),
\end{equation}
where $\bar{v}_\perp^{(l)}$ is the mean $v_\perp$ direction at layer $l$ and $\lambda \in [0, 1]$ controls the diversity-discriminability trade-off. At each greedy step the highest-scoring layer is added to $S$; the process repeats until $|S| = K$, yielding the final selected set $\mathcal{L}_{sel}$.

\subsubsection{Neuron selection via per-dimension Fisher scores}

Within each selected layer $l$, we further compress representations by identifying the most
discriminative neuron subsets. For each dimension $j \in \{1, \dots, d\}$, the 1D Fisher score is:
\begin{equation}
    f_j = \frac{(\mu_{1,j} - \mu_{0,j})^2}{\sigma_{1,j}^2 + \sigma_{0,j}^2 + \epsilon}.
\end{equation}
Neurons are ranked by $f_j$ and selected via an $\alpha$-threshold cumulative importance criterion:
the smallest set of neurons whose cumulative $f_j$ covers $\geq \alpha$ fraction of the total
Fisher importance is retained. This selection is applied independently to $h_Q^{(l)}$ and
$v_\perp^{(l)}$, respecting the geometric orthogonal decomposition and avoiding conflation of
contextual priors with deviation signals. The entire selection procedure requires only a single pass over training features, accumulating class-wise statistics for closed-form ranking.

\subsection{Joint feature construction}

The Fisher selection procedure (Section~\ref{sec:fisher_selection}) identifies the most discriminative neuron subsets from both $h_Q$ and $v_\perp$. Depending on the deployment scenario, these two feature sources can be combined in different ways: incorporating $h_Q$ provides explicit question context that can enhance in-domain discriminability, while using $v_\perp$ alone yields a more domain-agnostic probe. We therefore define two probe inputs that serve these different operational priorities.

\textbf{Question-orthogonal joint ($\mathcal{H}_{Q \oplus V_{\perp}}$)}: Concatenates question and question-orthogonal-component features, preserving both query context and deviation signal:
\begin{equation}
    \mathcal{H}_{Q \oplus V_{\perp}} = \mathrm{Concat}\left( [h_Q^{(l)}[\mathcal{N}_Q^{(l)}],
    v_{\perp}^{(l)}[\mathcal{N}_{v}^{(l)}]] \right)_{l \in \mathcal{L}_{sel}},
\end{equation}
where $\mathcal{N}_Q^{(l)}$ and $\mathcal{N}_v^{(l)}$ denote the Fisher-selected neuron
indices for $h_Q$ and $v_\perp$ at layer $l$, respectively.

\textbf{Question-orthogonal component ($\mathcal{H}_{V_{\perp}}$)}: Uses only the question-orthogonal component:
\begin{equation}
    \mathcal{H}_{V_{\perp}} = \mathrm{Concat}\left( v_{\perp}^{(l)}[\mathcal{N}_v^{(l)}]
    \right)_{l \in \mathcal{L}_{sel}}.
\end{equation}

The concatenated feature vector is standardized and fed into a lightweight MLP (1024 and 128 dimensions, ReLU, dropout, cross-entropy loss).

\section{Experiments}

\subsection{Experimental setup}

\textbf{Benchmarks.} We evaluate on four benchmarks: TriviaQA \citep{joshi-etal-2017-triviaqa}, SQuAD \citep{rajpurkar2018knowdontknowunanswerable}, NQ (Natural Questions) \citep{NQ}, and BioASQ \citep{Krithara2022.12.14.520213}.

\textbf{Models.} We test on four open-weight LLMs: gemma-2-2b \citep{gemmateam2024gemma2improvingopen}, Llama-2-7B-chat \citep{touvron2023llama2openfoundation}, Qwen3-14B \citep{yang2025qwen3technicalreport}, and Qwen3-30B-A3B \citep{yang2025qwen3technicalreport}.

\textbf{Baselines.} We compare against output-probability heuristics (PPL, Logit Entropy), black-box sampling consistency methods (SINdex \citep{sindex}, $P_{false}$ \citep{languagemodelsmostlyknow}), and internal state-based white-box probing methods ($P_{ik}$ \citep{languagemodelsmostlyknow}, MIND \citep{MIND}, SAPLMA \citep{SAPLMA}, and LLM-Check \citep{LLM-Check}). For SINdex, we follow the original setting with $N=10$ samples and \texttt{temperature}=$1.0$; other implementation details are provided in the appendix.

\textbf{Training configuration.} The \textsc{QAoD} probe is a two-hidden-layer MLP ($d_\text{in} \to 1024 \to 128 \to 2$, ReLU, dropout $0.1$). Class-weighted cross-entropy is used as the loss function to handle label imbalance, with weights inversely proportional to class frequency. The model is trained for 30 epochs with Adam ($\eta=0.001$, weight decay $0.01$).

\subsection{In-domain detection performance}

Table~\ref{tab:main_results} reports AUROC and F1 for both \textsc{QAoD} configurations against all baselines. Two patterns stand out:

\textbf{$\mathcal{H}_{Q \oplus V_{\perp}}$ achieves the best in-domain performance}, ranking first across all 16 model-dataset combinations by jointly encoding question context alongside the orthogonal deviation signal. Its advantage over the strongest single-pass baseline (SAPLMA) is consistent across all four model families, with gains of approximately 5--7 AUROC points on TriviaQA, indicating that the benefit of incorporating question context is robust to model scale and architecture. Notably, $\mathcal{H}_{Q \oplus V_\perp}$ consistently matches or surpasses SINdex across all model-dataset pairs, yet requires only a single forward pass rather than ten. This suggests that internal geometric structure encodes discriminative information at least as rich as that extracted through repeated output sampling.

\textbf{$\mathcal{H}_{V_{\perp}}$ remains competitive in-domain} while consistently leading zero-shot OOD evaluation in Table~\ref{tab:ood_results}. Despite using no question context, it matches or exceeds SAPLMA on three of the four models in-domain, confirming that the orthogonal decomposition alone captures strong factuality signal and making it the preferred choice when cross-domain generalization takes priority.

\begin{table}[!ht]
    \centering
    \renewcommand{\arraystretch}{0.99}
    \setlength{\tabcolsep}{7pt}
    \caption{Hallucination detection performance on four datasets (seed=42), reporting AUROC with F1 in parentheses (\%). \textbf{Bold}: best; \underline{underline}: second best. $\text{\textsc{QAoD}}_{Q \oplus V_\perp}$ ranks first on all 16 model-dataset cells.}
    \label{tab:main_results}
    \small
    \begin{tabular}{llcccc}
    \toprule
    \textbf{Model} & \textbf{Method} & \textbf{TriviaQA} & \textbf{SQuAD} & \textbf{NQ} & \textbf{BioASQ} \\
    \cmidrule(lr){3-6}
    & & \footnotesize{AUC (F1)} & \footnotesize{AUC (F1)} & \footnotesize{AUC (F1)} & \footnotesize{AUC (F1)} \\
    \midrule
    \multirow{10}{*}{gemma-2-2b}
    & PPL & 73.90 {\scriptsize(38.28)} & 76.18 {\scriptsize(40.34)} & 67.48 {\scriptsize(84.22)} & 59.32 {\scriptsize(69.82)} \\
    & Logit Entropy & 75.59 {\scriptsize(38.94)} & 77.35 {\scriptsize(40.54)} & 69.14 {\scriptsize(84.44)} & 64.80 {\scriptsize(69.89)} \\
    & SINdex & 80.19 {\scriptsize(42.38)} & 78.26 {\scriptsize(42.00)} & 75.20 {\scriptsize(85.05)} & 72.91 {\scriptsize(71.46)} \\
    & $P_{false}$ & 69.51 {\scriptsize(35.12)} & 65.29 {\scriptsize(33.65)} & 69.89 {\scriptsize(83.73)} & 65.45 {\scriptsize(72.34)} \\
    & LLM-Check & 59.97 {\scriptsize(27.62)} & 59.18 {\scriptsize(28.13)} & 49.90 {\scriptsize(83.63)} & 63.99 {\scriptsize(69.14)} \\
    & $P_{ik}$ & 77.12 {\scriptsize(41.62)} & 69.31 {\scriptsize(35.16)} & 71.58 {\scriptsize(83.72)} & 72.51 {\scriptsize(70.35)} \\
    
    & MIND & 76.80 {\scriptsize(49.94)} & 66.76 {\scriptsize(34.20)} & 72.41 {\scriptsize(84.00)} & 73.42 {\scriptsize(70.94)} \\
    & SAPLMA & \underline{83.32} {\scriptsize(51.42)} & 77.23 {\scriptsize(42.29)} & \underline{78.40} {\scriptsize(84.96)} & \underline{75.66} {\scriptsize(71.06)} \\
    
    \cmidrule{2-6}
    & $\text{\textsc{QAoD}}_{V_{\perp}}$ & 82.26 {\scriptsize(52.18)} & \underline{79.73} {\scriptsize(47.55)} & 77.91 {\scriptsize(85.19)} & 74.64 {\scriptsize(70.93)} \\
    & $\text{\textsc{QAoD}}_{Q \oplus V_{\perp}}$ & \textbf{90.10} {\scriptsize(63.75)} & \textbf{82.16} {\scriptsize(48.84)} & \textbf{81.42} {\scriptsize(85.68)} & \textbf{81.10} {\scriptsize(74.70)} \\
    \midrule
    \multirow{10}{*}{Llama-2-7B-chat}
    & PPL & 52.16 {\scriptsize(23.59)} & 56.19 {\scriptsize(22.59)} & 56.81 {\scriptsize(78.94)} & 53.39 {\scriptsize(64.23)} \\
    & Logit Entropy & 52.23 {\scriptsize(23.59)} & 56.48 {\scriptsize(22.59)} & 56.85 {\scriptsize(78.94)} & 57.10 {\scriptsize(64.23)} \\
    & SINdex & 68.93 {\scriptsize(40.65)} & 75.72 {\scriptsize(39.56)} & 73.78 {\scriptsize(79.72)} & 79.19 {\scriptsize(68.44)} \\
    & $P_{false}$ & 65.12 {\scriptsize(31.29)} & 69.42 {\scriptsize(32.00)} & 68.08 {\scriptsize(79.57)} & 73.76 {\scriptsize(68.71)} \\
    & LLM-Check & 65.06 {\scriptsize(30.23)} & 57.35 {\scriptsize(25.50)} & 56.06 {\scriptsize(79.14)} & 71.40 {\scriptsize(71.13)} \\
    & $P_{ik}$ & 75.40 {\scriptsize(43.63)} & 65.93 {\scriptsize(37.26)} & 62.37 {\scriptsize(79.30)} & 72.48 {\scriptsize(67.62)} \\
    
    & MIND & 79.31 {\scriptsize(47.32)} & 68.06 {\scriptsize(36.10)} & 67.10 {\scriptsize(79.72)} & 76.81 {\scriptsize(72.66)} \\
    & SAPLMA & 84.33 {\scriptsize(52.74)} & 79.47 {\scriptsize(43.25)} & 74.00 {\scriptsize(80.90)} & \underline{79.21} {\scriptsize(69.35)} \\
    
    \cmidrule{2-6}
    & $\text{\textsc{QAoD}}_{V_{\perp}}$ & \underline{86.20} {\scriptsize(54.26)} & \underline{80.75} {\scriptsize(44.08)} & \underline{75.08} {\scriptsize(81.20)} & 77.46 {\scriptsize(74.32)} \\
    & $\text{\textsc{QAoD}}_{Q \oplus V_{\perp}}$ & \textbf{90.21} {\scriptsize(61.28)} & \textbf{83.09} {\scriptsize(47.47)} & \textbf{77.94} {\scriptsize(81.68)} & \textbf{83.45} {\scriptsize(76.08)} \\
    \midrule
    \multirow{10}{*}{Qwen3-14B}
    & PPL & 53.25 {\scriptsize(13.13)} & 54.89 {\scriptsize(14.75)} & 61.81 {\scriptsize(49.47)} & 61.33 {\scriptsize(46.28)} \\
    & Logit Entropy & 53.25 {\scriptsize(12.94)} & 54.92 {\scriptsize(14.65)} & 62.06 {\scriptsize(49.47)} & 61.88 {\scriptsize(46.52)} \\
    & SINdex & 65.73 {\scriptsize(32.08)} & 74.02 {\scriptsize(20.46)} & 78.30 {\scriptsize(74.65)} & 80.50 {\scriptsize(67.49)} \\
    & $P_{false}$ & 78.04 {\scriptsize(32.71)} & 58.33 {\scriptsize(12.33)} & 79.05 {\scriptsize(77.71)} & 79.17 {\scriptsize(66.97)} \\
    & LLM-Check & 61.32 {\scriptsize(21.81)} & 62.64 {\scriptsize(13.06)} & 53.16 {\scriptsize(77.03)} & 59.67 {\scriptsize(57.42)} \\
    & $P_{ik}$ & 79.14 {\scriptsize(39.61)} & 69.76 {\scriptsize(19.45)} & 67.73 {\scriptsize(70.52)} & 73.22 {\scriptsize(64.13)} \\
    & MIND & 81.35 {\scriptsize(45.99)} & 72.12 {\scriptsize(24.03)} & 73.69 {\scriptsize(78.34)} & 76.81 {\scriptsize(65.41)} \\
    & SAPLMA & 85.45 {\scriptsize(54.26)} & \underline{83.57} {\scriptsize(29.03)} & 78.93 {\scriptsize(79.03)} & \underline{81.66} {\scriptsize(68.59)} \\
    
    \cmidrule{2-6}
    & $\text{\textsc{QAoD}}_{V_{\perp}}$ & \underline{87.42} {\scriptsize(58.67)} & 81.34 {\scriptsize(29.39)} & \underline{79.98} {\scriptsize(80.25)} & 78.39 {\scriptsize(67.14)} \\
    & $\text{\textsc{QAoD}}_{Q \oplus V_{\perp}}$ & \textbf{90.13} {\scriptsize(60.81)} & \textbf{85.01} {\scriptsize(32.09)} & \textbf{82.69} {\scriptsize(81.25)} & \textbf{85.69} {\scriptsize(73.51)} \\
    \midrule
    \multirow{10}{*}{Qwen3-30B-A3B}
    & PPL & 54.51 {\scriptsize(16.85)} & 55.78 {\scriptsize(17.74)} & 60.33 {\scriptsize(77.54)} & 60.07 {\scriptsize(52.34)} \\
    & Logit Entropy & 54.52 {\scriptsize(16.85)} & 55.82 {\scriptsize(17.74)} & 60.47 {\scriptsize(77.54)} & 59.73 {\scriptsize(52.34)} \\
    & SINdex & 68.26 {\scriptsize(36.68)} & 74.07 {\scriptsize(35.46)} & 77.33 {\scriptsize(79.43)} & 81.37 {\scriptsize(67.67)} \\
    & $P_{false}$ & 80.90 {\scriptsize(44.52)} & 70.59 {\scriptsize(22.79)} & 80.30 {\scriptsize(81.84)} & 79.33 {\scriptsize(66.81)} \\
    & LLM-Check & 66.89 {\scriptsize(25.95)} & 58.69 {\scriptsize(18.21)} & 55.53 {\scriptsize(77.52)} & 62.25 {\scriptsize(55.22)} \\
    & $P_{ik}$ & 74.04 {\scriptsize(42.11)} & 71.60 {\scriptsize(27.71)} & 69.99 {\scriptsize(77.61)} & 70.32 {\scriptsize(59.20)} \\
    & MIND & 81.38 {\scriptsize(49.57)} & 72.90 {\scriptsize(30.48)} & 72.69 {\scriptsize(78.30)} & 77.87 {\scriptsize(65.78)} \\
    & SAPLMA & 86.74 {\scriptsize(52.23)} & \underline{83.31} {\scriptsize(39.76)} & 78.78 {\scriptsize(80.27)} & \underline{83.24} {\scriptsize(68.22)} \\
    
    \cmidrule{2-6}
    & $\text{\textsc{QAoD}}_{V_{\perp}}$ & \underline{87.26} {\scriptsize(53.90)} & 82.52 {\scriptsize(36.45)} & \underline{79.27} {\scriptsize(80.48)} & 79.95 {\scriptsize(67.85)} \\
    & $\text{\textsc{QAoD}}_{Q \oplus V_{\perp}}$ & \textbf{91.84} {\scriptsize(63.89)} & \textbf{85.50} {\scriptsize(40.81)} & \textbf{82.76} {\scriptsize(81.76)} & \textbf{86.04} {\scriptsize(73.38)} \\
    \midrule
    \end{tabular}
    \end{table}

\subsection{Cross-domain generalization}
\label{sec:ood}

To evaluate cross-domain transferability, we train detectors on a general-domain mixture of TriviaQA, SQuAD, and NQ, and evaluate them directly on BioASQ without any target-domain fine-tuning. The BioASQ training split is used only for the in-domain results, while the zero-shot OOD setting is reserved for cross-domain evaluation.

\begin{table}[!ht]
    \centering
    \caption{Zero-shot cross-domain generalization to BioASQ (AUROC, \%, seed=42). Detectors trained on \{TriviaQA, SQuAD, NQ\} are evaluated directly on BioASQ; \textbf{bold}/\underline{underline}: best/second best. $\mathcal{H}_{V_\perp}$ leads on all four models by a substantial margin.}
    \label{tab:ood_results}
    \small
    \begin{tabular}{lccccc}
    \toprule
    \textbf{Model} & $P_{ik}$ & MIND & SAPLMA
      & $\text{\textsc{QAoD}}_{V_{\perp}}$ & $\text{\textsc{QAoD}}_{Q \oplus V_{\perp}}$ \\
    \midrule
    gemma-2-2b
      & 45.10 & 48.35 & 59.15
      & \textbf{72.19} & \underline{70.95} \\
    Llama-2-7B-chat
      & 55.72 & 45.77 & 56.76
      & \textbf{75.63} & \underline{65.28} \\
    Qwen3-14B
      & 58.48 & 51.99 & 55.29
      & \textbf{76.73} & \underline{71.62} \\
    Qwen3-30B-A3B
      & 66.10 & 62.95 & 63.63
      & \textbf{77.51} & \underline{74.27} \\
    
    \bottomrule
    \end{tabular}
    \end{table}

\begin{figure}[!ht]
\centering
\begin{subfigure}[t]{0.48\textwidth}
    \centering
    \includegraphics[width=\linewidth]{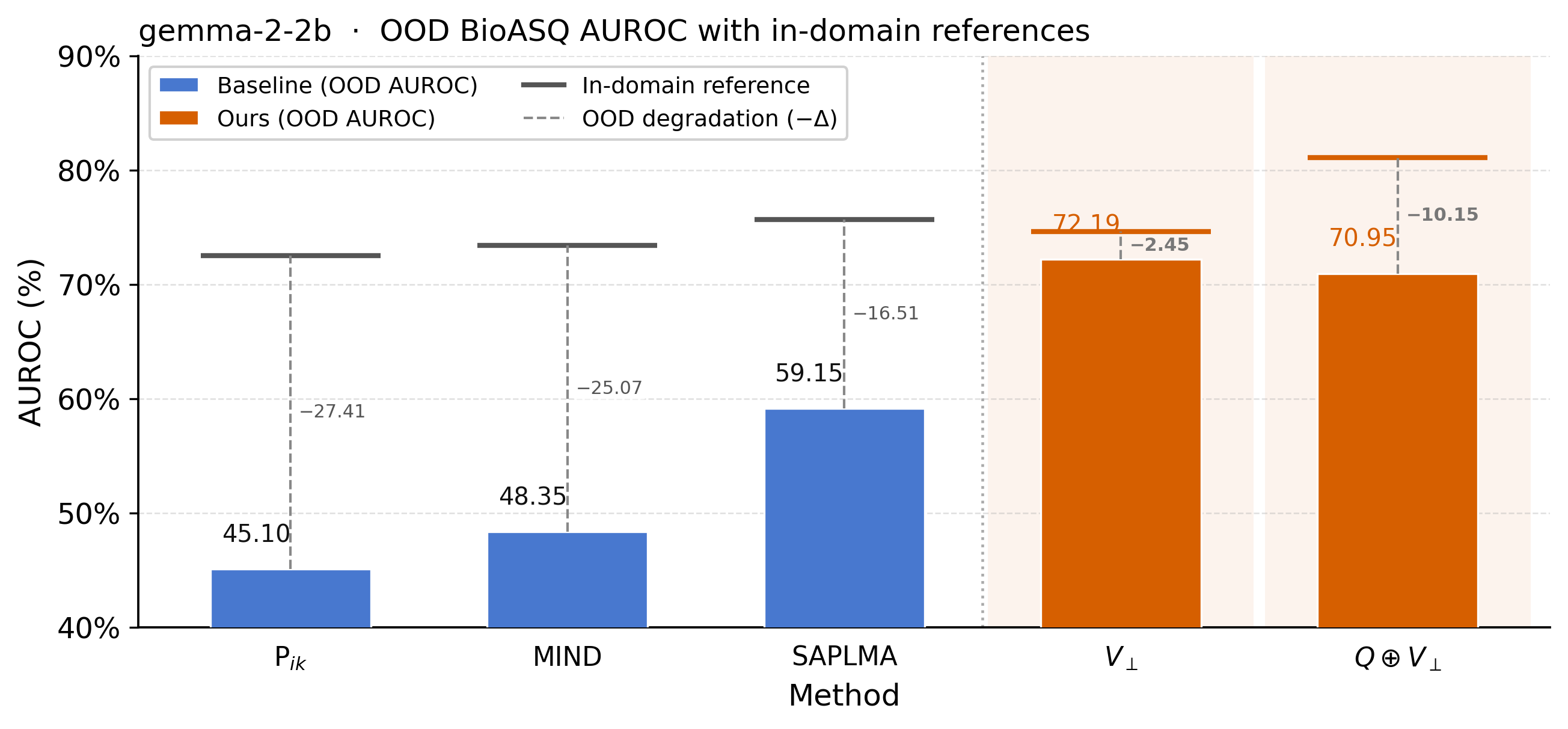}
    \caption{gemma-2-2b}
\end{subfigure}
\hfill
\begin{subfigure}[t]{0.48\textwidth}
    \centering
    \includegraphics[width=\linewidth]{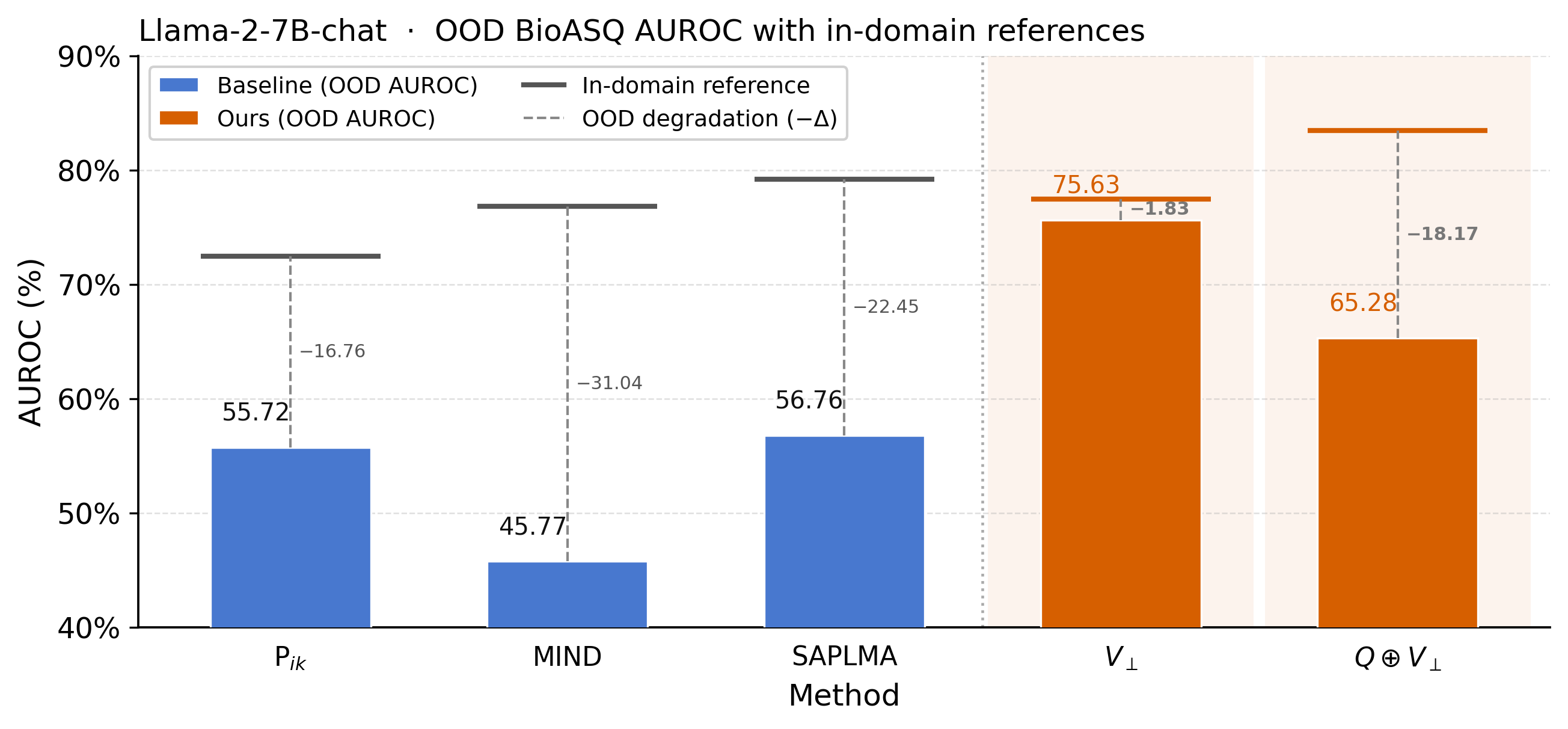}
    \caption{Llama-2-7B-chat}
\end{subfigure}

\vspace{0.5em}

\begin{subfigure}[t]{0.48\textwidth}
    \centering
    \includegraphics[width=\linewidth]{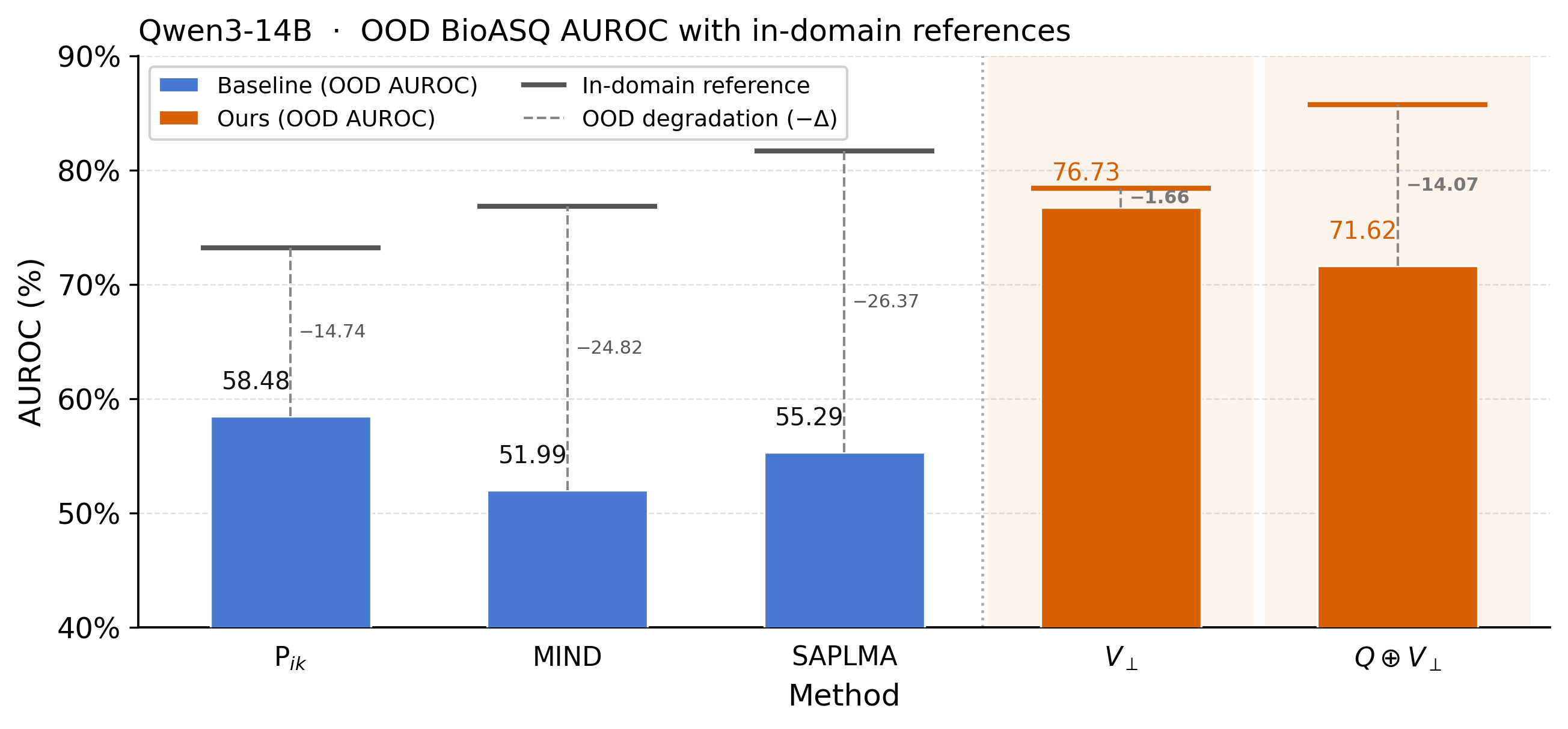}
    \caption{Qwen3-14B}
\end{subfigure}
\hfill
\begin{subfigure}[t]{0.48\textwidth}
    \centering
    \includegraphics[width=\linewidth]{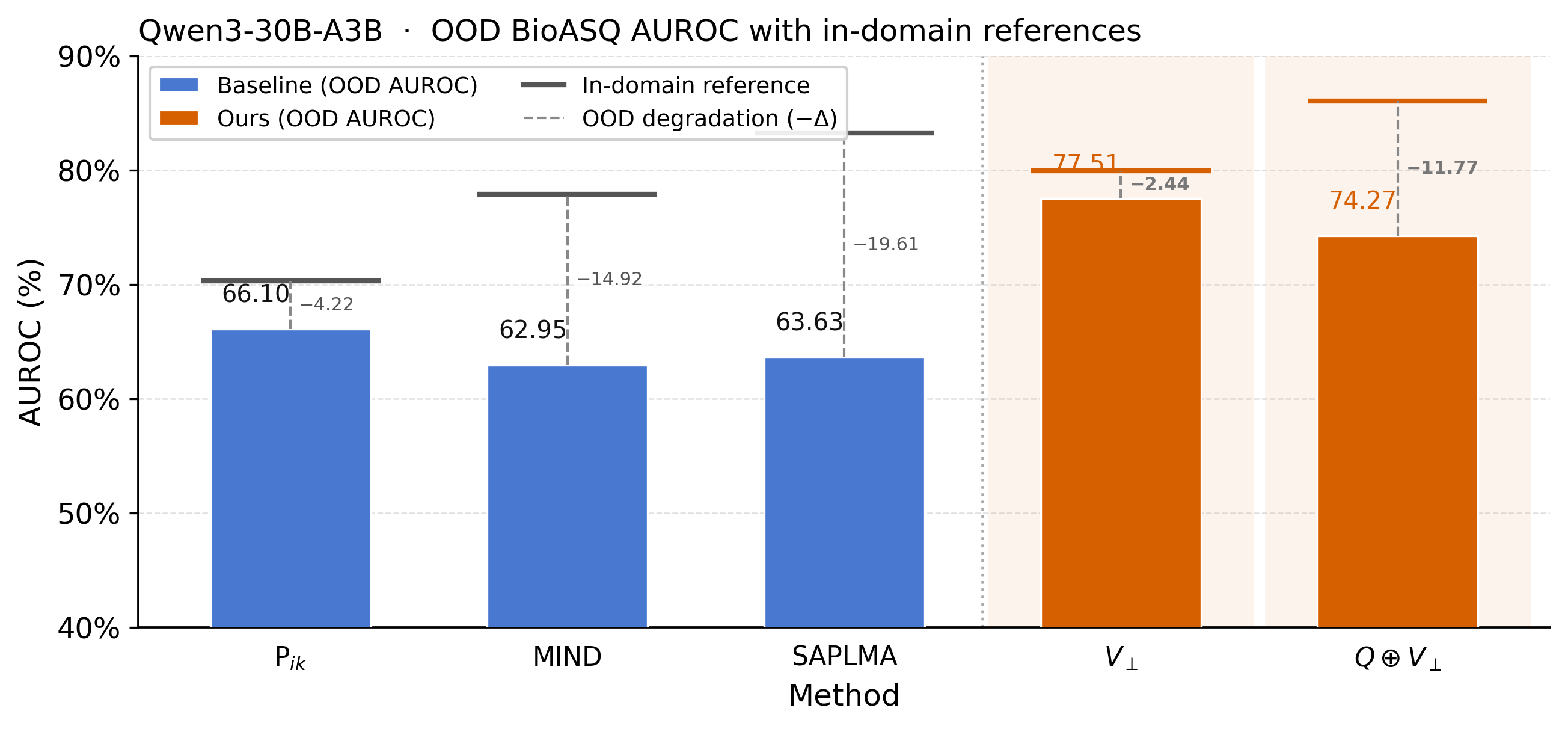}
    \caption{Qwen3-30B-A3B}
\end{subfigure}

\caption{Zero-shot OOD AUROC on BioASQ across four LLMs. Bars show OOD AUROC for detectors trained on \{TriviaQA, SQuAD, NQ\} and evaluated on BioASQ; the tick marks indicate in-domain BioASQ AUROC and the dashed line the degradation gap. $\text{\textsc{QAoD}}_{V_\perp}$ achieves the best OOD AUROC and smallest gap.}
\label{fig:ood_bars}
\end{figure}

\label{sec:ood_mechanism}

\begin{figure}[!ht]
    \centering
    \begin{subfigure}[t]{0.48\textwidth}
        \centering
        \includegraphics[width=\linewidth]{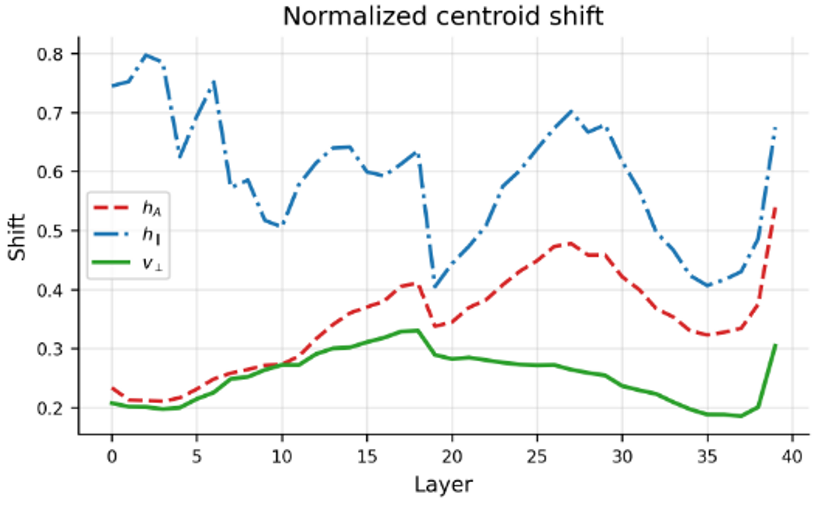}
        \caption{Qwen3-14B}
    \end{subfigure}
    \hfill
    \begin{subfigure}[t]{0.48\textwidth}
        \centering
        \includegraphics[width=\linewidth]{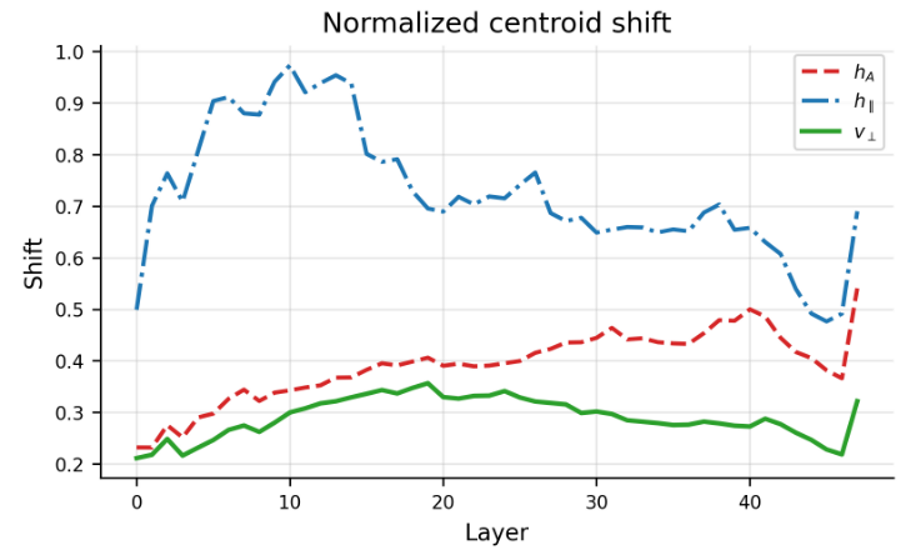}
        \caption{Qwen3-30B-A3B}
    \end{subfigure}
    \caption{Layer-wise normalized centroid shift ($\|\mu_{\text{src}} - \mu_{\text{tgt}}\| / \bar{\sigma}$) from \{TriviaQA, SQuAD, NQ\} to BioASQ. The question-orthogonal component $v_\perp$ exhibits the smallest displacement across all layers, with domain-dependent variation concentrated in the question-aligned component $h_\parallel$.}
    \label{fig:domain_drift}
\end{figure}

Cross-domain generalization to BioASQ is the primary distinguishing capability of \textsc{QAoD} relative to existing white-box detectors. As shown in Table~\ref{tab:ood_results} and Figure~\ref{fig:ood_bars}, $\mathcal{H}_{V_\perp}$ achieves the highest zero-shot BioASQ AUROC across all four models, surpassing the strongest answer-side baseline (SAPLMA) by more than 10 AUROC points on all four models and up to 21 AUROC points, while answer-side probes degrade sharply under the source-to-target domain shift. We analyze the geometric mechanism underlying this advantage on Qwen3-14B and Qwen3-30B-A3B, which serve as representative dense and MoE backbones covering two qualitatively different architectures.

\paragraph{Geometric view.}
Figure~\ref{fig:domain_drift} plots the normalized centroid shift
$\|\mu_{\text{src}} - \mu_{\text{tgt}}\| / \bar{\sigma}$ from the
source-domain mixture to BioASQ. For both representative models, the same
layer-wise ordering is observed: the question-aligned component
$h_\parallel = \frac{h_A^{(l)} \cdot h_Q^{(l)}}{\|h_Q^{(l)}\|^2} h_Q^{(l)}$ undergoes the largest displacement, the original answer state
$h_A$ lies in the middle, and the question-orthogonal component $v_\perp$ is the least
displaced. Much of the domain-dependent variation thus aligns with the question direction; subtracting it preserves the remaining answer information less tied to domain change.

\paragraph{CKA evidence.}
We further use linear CKA \citep{kornblith2019similarityneuralnetworkrepresentations} to characterize the reduction of domain-related
variation from an information-alignment perspective. Specifically,
$\mathrm{CKA}_\text{domain}$ measures linear alignment with domain identity
(\{TriviaQA, SQuAD, NQ\} vs.\ BioASQ). Table~\ref{tab:cka_ratio} shows a consistent pattern on
both Qwen3-14B and Qwen3-30B-A3B: $v_\perp$ has the lowest domain alignment
across all three feature-selection regimes, confirming that the question-orthogonal
component suppresses domain-identity information more effectively than $h_A$
or $h_\parallel$. The CKA result is therefore complementary to the centroid-shift analysis; the two converge on the same conclusion: $v_\perp$ is less sensitive to the source-target domain change.

Importantly, $\mathrm{CKA}_\text{hall}$ for $v_\perp$ is comparable to $h_A$
in the all-layer setting and surpasses it after Fisher selection, confirming
that the reduction in domain alignment does not sacrifice hallucination discriminability.
The selectivity ratio $\mathrm{CKA}_\text{hall}/\mathrm{CKA}_\text{domain}$ captures this joint property: $v_\perp$ achieves the highest ratio under every selection regime (Table~\ref{tab:cka_ratio}), and Fisher selection amplifies this advantage while decreasing the ratio for $h_A$ and $h_\parallel$, confirming that their discriminative dimensions are more entangled with domain identity.

\begin{table}[!ht]
\centering
\caption{Linear CKA alignment with hallucination ($\mathrm{CKA}_\text{hall}$) and domain-membership ($\mathrm{CKA}_\text{domain}$) labels, and selectivity ratio, at three feature-selection operating points; $v_\perp$ achieves the highest ratio under every regime, and Fisher selection widens the advantage.}
\label{tab:cka_ratio}
\scriptsize
\setlength{\tabcolsep}{3.5pt}
\resizebox{\textwidth}{!}{%
\begin{tabular}{ll ccc ccc ccc}
\toprule
& & \multicolumn{3}{c}{$\mathrm{CKA}_\text{hall}$}
  & \multicolumn{3}{c}{$\mathrm{CKA}_\text{domain}$}
  & \multicolumn{3}{c}{$\mathrm{CKA}_\text{hall}/\mathrm{CKA}_\text{domain}$} \\
\cmidrule(lr){3-5}\cmidrule(lr){6-8}\cmidrule(lr){9-11}
\textbf{Model} & \textbf{Selection}
  & $h_A$ & $h_\parallel$ & $v_\perp$
  & $h_A$ & $h_\parallel$ & $v_\perp$
  & $h_A$ & $h_\parallel$ & $v_\perp$ \\
\midrule
\multirow{3}{*}{Qwen3-14B}
  & All layers
    & 0.099 & 0.090 & \textbf{0.068}
    & 0.207 & 0.155 & \textbf{0.114}
    & 0.476 & 0.584 & \textbf{0.596} \\
  & Fisher layer
    & 0.073 & 0.068 & \textbf{0.072}
    & 0.218 & 0.171 & \textbf{0.115}
    & 0.334 & 0.398 & \textbf{0.621} \\
  & Fisher layer+neuron
    & 0.069 & 0.065 & \textbf{0.072}
    & 0.202 & 0.151 & \textbf{0.109}
    & 0.344 & 0.434 & \textbf{0.663} \\
\midrule
\multirow{3}{*}{Qwen3-30B-A3B}
  & All layers
    & 0.083 & 0.077 & \textbf{0.080}
    & 0.269 & 0.226 & \textbf{0.149}
    & 0.310 & 0.343 & \textbf{0.537} \\
  & Fisher layer
    & 0.059 & 0.046 & \textbf{0.076}
    & 0.240 & 0.196 & \textbf{0.137}
    & 0.244 & 0.237 & \textbf{0.556} \\
  & Fisher layer+neuron
    & 0.056 & 0.044 & \textbf{0.077}
    & 0.236 & 0.184 & \textbf{0.136}
    & 0.239 & 0.238 & \textbf{0.564} \\
\bottomrule
\end{tabular}%
}
\end{table}

\subsection{Ablation studies}
\label{sec:ablation_studies}

We report three ablations on Qwen3-14B and Qwen3-30B-A3B (dense and MoE representatives): a zero-shot OOD feature-construction study (trained on \{TriviaQA, SQuAD, NQ\}, tested on BioASQ), a layer-selection comparison on Qwen3-14B / TriviaQA, and a neuron-threshold sweep. All other settings follow the main experiments.

\begin{figure}[!ht]
\centering
\begin{minipage}[t]{0.37\textwidth}
\centering
\small
\setlength{\tabcolsep}{2.0pt}
\makebox[\linewidth][c]{%
\scalebox{0.92}[1]{%
\begin{tabular}{@{}lcc@{}}
\toprule
\textbf{Feature} & \textbf{14B} & \textbf{30B} \\
\midrule
Random & 64.24 $\pm$ 1.57 & 72.58 $\pm$ 1.66 \\
Q-only & 71.12 & 73.03 \\
A-only & 66.09 & 72.35 \\
Q+A (no proj.) & 69.88 & 76.11 \\
\textsc{QAoD} ($\mathcal{H}_{V_\perp}$) & \textbf{76.73} & \textbf{77.51} \\
\bottomrule
\end{tabular}%
}}
\caption{Feature-construction ablation: zero-shot OOD AUROC (\%) on BioASQ; correct orthogonalization is necessary.}
\label{tab:feature_ablation}
\end{minipage}
\hfill
\begin{minipage}[t]{0.61\textwidth}
\centering
\small
\setlength{\tabcolsep}{1.2pt}
\makebox[\linewidth][c]{%
\scalebox{0.82}[1]{%
\begin{tabular}{@{}lccc c@{}}
\toprule
\textbf{Method} & \textbf{$K=5$} & \textbf{$K=10$} & \textbf{$K=15$} & \textbf{Layers ($K{=}5$)} \\
\midrule
Last-$N$ layers & 84.34 & 85.06 & 86.64 & [35,36,37,38,39] \\
Pure Fisher & 84.92 & 86.02 & 86.42 & [19,20,21,22,23] \\
Uniform spacing & 86.83 & 87.03 & 87.02 & [0,10,20,29,39] \\
Random & 86.49$\pm$0.43 & 86.74$\pm$0.19 & 86.86$\pm$0.37 & --- \\
DPF (ours) & \textbf{87.27} & \textbf{87.36} & \textbf{87.42} & [10,19,26,33,39] \\
\bottomrule
\end{tabular}%
}}
\caption{Layer-selection ablation on Qwen3-14B / TriviaQA (AUROC, \%); DPF leads at every budget and outperforms full-layer aggregation.}
\label{tab:layer_ablation}
\end{minipage}
\end{figure}

Table~\ref{tab:feature_ablation} holds Fisher selection fixed and varies only the feature combination. The three non-\textsc{QAoD} baselines all omit orthogonalization: \textit{Random} projects the answer onto a random unit vector instead of the question direction; \textit{Q-only} and \textit{A-only} apply Fisher selection to $h_Q$ or $h_A$ alone; and \textit{Q+A (no proj.)} follows the same Fisher selection and concatenation pipeline as \textsc{QAoD} but concatenates $h_Q$ and $h_A$ directly without computing $v_\perp$. \textit{Random} performs worst, confirming that OOD gains require geometrically meaningful decoupling; \textit{Q-only}/\textit{A-only} each underperform \textsc{QAoD}, and \textit{Q+A (no proj.)} also fails to match it, showing the advantage depends on correct question-aware orthogonalization rather than mere concatenation.

In Table~\ref{tab:layer_ablation}, combining discriminability with diversity is the key factor: \textit{Last-$N$ layers} and pure top-$K$ \textit{Fisher} both select largely redundant consecutive layers, limiting coverage of complementary processing stages. \textit{Uniform spacing} and \textit{Random} selection improve over these by increasing layer diversity, but without discriminability guidance. DPF combines both objectives, consistently achieving the best AUROC at every budget and selecting well-spread layers (e.g., $[10,19,26,33,39]$ for $K{=}5$). Using all 40 layers (AUROC 86.95) also falls below DPF at $K{=}15$ (87.42), confirming that diversity-aware compact selection outperforms full-depth aggregation.

\begin{figure}[!ht]
\centering
\includegraphics[width=0.95\linewidth]{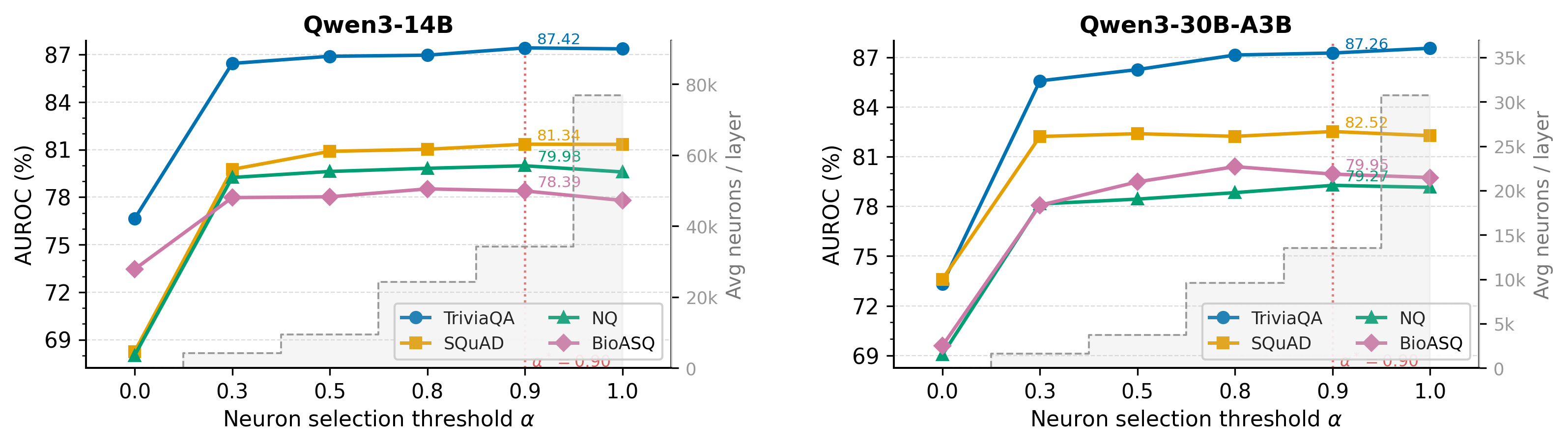}
\caption{Neuron-selection ablation on the cumulative Fisher threshold $\alpha$ under $K=15$ and $\lambda=1$; performance stabilizes at $\alpha \in [0.8, 0.9]$, motivating the default $\alpha=0.9$.}
\label{fig:alpha_ablation}
\end{figure}

Figure~\ref{fig:alpha_ablation} shows performance saturating around $\alpha\in[0.8,0.9]$; we adopt $\alpha=0.9$ as the default.

\subsection{Efficiency}
\label{sec:efficiency}

\begin{figure}[!ht]
\centering
\includegraphics[width=0.9\linewidth]{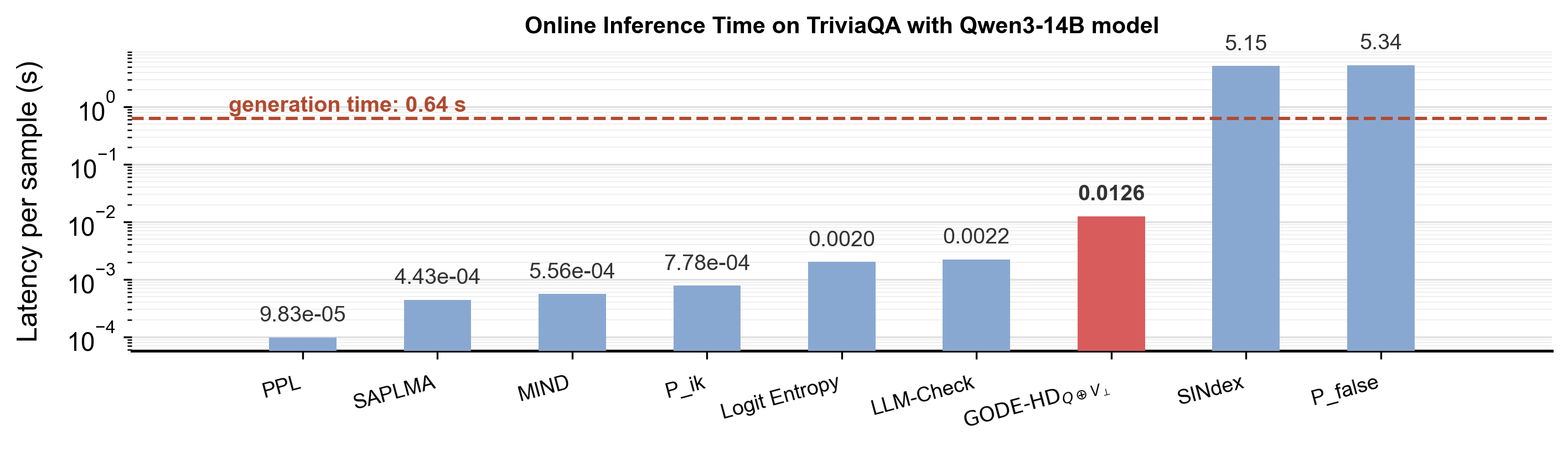}
\caption{Inference-time comparison (log scale), including shared generation cost and additional detection overhead; \textsc{QAoD}'s detection cost is negligible (${\approx}2\%$) relative to generation time.}
\label{fig:latency}
\end{figure}

Because all methods share the same answer-generation forward pass, we compare only the additional detection overhead. Figure~\ref{fig:latency} shows the runtime breakdown on TriviaQA with Qwen3-14B (NVIDIA A100, batch size 256), highlighting the substantial computational gap between repeated-sampling methods and single-pass white-box probes in practice.

Detection overhead is orders of magnitude lower than sampling-based methods such as SINdex ($N{=}10$). Among single-pass white-box probes, \textsc{QAoD}'s per-query detection overhead of $0.0126$s is marginally higher than lighter baselines, but remains negligible relative to the shared generation cost of ${\sim}0.64$s (${\approx}2\%$). Given the substantial OOD robustness advantage demonstrated in Section~\ref{sec:ood}, this marginal overhead is well-justified within the single-pass operating regime.

\section{Limitations}
The current evaluation covers knowledge-intensive QA tasks, where factual errors can often be traced to specific misapplied entities or relationships. Extending \textsc{QAoD} to tasks such as mathematical reasoning or open-ended generation, where the geometric properties of hidden states may differ substantially, is a promising direction for future work. Additionally, \textsc{QAoD} is inherently a white-box method: it requires access to intermediate hidden states of the target LLM. This makes it not directly applicable to closed-source models accessed only through inference APIs, which constitute a significant and growing fraction of deployed systems. Extending it to such settings, for instance by transferring probe knowledge from an accessible surrogate model to a black-box target, is a direction we plan to explore in future work.

\section{Conclusion}

In LLM hallucination detection, black-box methods demand repeated inference while white-box probes degrade sharply under domain shift. We proposed \textsc{QAoD}, which resolves this tension by orthogonally projecting out the question-aligned component from answer representations and applying Fisher-discriminant scoring to select compact, domain-stable probes. This is grounded in the geometric observation that domain-sensitive variation concentrates in the question-aligned direction and the residual question-orthogonal component suppresses domain-conditioned noise while retaining factuality signals. Evaluated across four LLMs and four benchmarks, $\mathcal{H}_{Q \oplus V_\perp}$ achieves the best in-domain AUROC on all model-dataset pairs, while $\mathcal{H}_{V_\perp}$ surpasses the best white-box baseline by up to 21\% points on zero-shot BioASQ. These results suggest that explicitly modeling the geometric relationship between question and answer representations is a promising direction toward hallucination detectors that are simultaneously efficient and robust to domain shift, particularly in real-world settings where target-domain labels are scarce.

\bibliographystyle{unsrtnat}
\bibliography{references}

@misc{selfcheckgpt,
      title={SelfCheckGPT: Zero-Resource Black-Box Hallucination Detection for Generative Large Language Models},
      author={Potsawee Manakul and Adian Liusie and Mark J. F. Gales},
      year={2023},
      eprint={2303.08896},
      archivePrefix={arXiv},
      primaryClass={cs.CL},
      url={https://arxiv.org/abs/2303.08896},
}

@article{semanticentropy,
  author  = {Farquhar, Sebastian and Kossen, Jannik and Kuhn, Lorenz and Gal, Yarin},
  title   = {Detecting hallucinations in large language models using semantic entropy},
  journal = {Nature},
  year    = {2024},
  volume  = {630},
  number  = {8017},
  pages   = {625--630},
  month   = {06},
  doi     = {10.1038/s41586-024-07421-0},
  url     = {https://doi.org/10.1038/s41586-024-07421-0}
}

@misc{languagemodelsmostlyknow,
      title={Language Models (Mostly) Know What They Know},
      author={Saurav Kadavath and Tom Conerly and Amanda Askell and Tom Henighan and Dawn Drain and Ethan Perez and Nicholas Schiefer and Zac Hatfield-Dodds and Nova DasSarma and Eli Tran-Johnson and Scott Johnston and Sheer El-Showk and Andy Jones and Nelson Elhage and Tristan Hume and Anna Chen and Yuntao Bai and Sam Bowman and Stanislav Fort and Deep Ganguli and Danny Hernandez and Josh Jacobson and Jackson Kernion and Shauna Kravec and Liane Lovitt and Kamal Ndousse and Catherine Olsson and Sam Ringer and Dario Amodei and Tom Brown and Jack Clark and Nicholas Joseph and Ben Mann and Sam McCandlish and Chris Olah and Jared Kaplan},
      year={2022},
      eprint={2207.05221},
      archivePrefix={arXiv},
      primaryClass={cs.CL},
      url={https://arxiv.org/abs/2207.05221},
}

@inproceedings{MIND,
    title = "Unsupervised Real-Time Hallucination Detection based on the Internal States of Large Language Models",
    author = "Su, Weihang  and
      Wang, Changyue  and
      Ai, Qingyao  and
      Hu, Yiran  and
      Wu, Zhijing  and
      Zhou, Yujia  and
      Liu, Yiqun",
    editor = "Ku, Lun-Wei  and
      Martins, Andre  and
      Srikumar, Vivek",
    booktitle = "Findings of the Association for Computational Linguistics: ACL 2024",
    month = aug,
    year = "2024",
    address = "Bangkok, Thailand",
    publisher = "Association for Computational Linguistics",
    url = "https://aclanthology.org/2024.findings-acl.854/",
    doi = "10.18653/v1/2024.findings-acl.854",
    pages = "14379--14391"
}

@misc{geometrytruthemergentlinear,
      title={The Geometry of Truth: Emergent Linear Structure in Large Language Model Representations of True/False Datasets},
      author={Samuel Marks and Max Tegmark},
      year={2024},
      eprint={2310.06824},
      archivePrefix={arXiv},
      primaryClass={cs.AI},
      url={https://arxiv.org/abs/2310.06824},
}

@misc{implicitrepresentationsmeaningneural,
      title={Implicit Representations of Meaning in Neural Language Models},
      author={Belinda Z. Li and Maxwell Nye and Jacob Andreas},
      year={2021},
      eprint={2106.00737},
      archivePrefix={arXiv},
      primaryClass={cs.CL},
      url={https://arxiv.org/abs/2106.00737},
}

@misc{SAPLMA,
      title={The Internal State of an LLM Knows When It's Lying},
      author={Amos Azaria and Tom Mitchell},
      year={2023},
      eprint={2304.13734},
      archivePrefix={arXiv},
      primaryClass={cs.CL},
      url={https://arxiv.org/abs/2304.13734},
}

@article{Levinstein_2024,
   title={Still no lie detector for language models: probing empirical and conceptual roadblocks},
   volume={182},
   ISSN={1573-0883},
   url={http://dx.doi.org/10.1007/s11098-023-02094-3},
   DOI={10.1007/s11098-023-02094-3},
   number={7},
   journal={Philosophical Studies},
   publisher={Springer Science and Business Media LLC},
   author={Levinstein, Benjamin A. and Herrmann, Daniel A.},
   year={2024},
   month=feb, pages={1539–1565} }

@misc{DoLa,
      title={DoLa: Decoding by Contrasting Layers Improves Factuality in Large Language Models},
      author={Yung-Sung Chuang and Yujia Xie and Hongyin Luo and Yoon Kim and James Glass and Pengcheng He},
      year={2024},
      eprint={2309.03883},
      archivePrefix={arXiv},
      primaryClass={cs.CL},
      url={https://arxiv.org/abs/2309.03883},
}

@misc{sindex,
      title={SINdex: Semantic INconsistency Index for Hallucination Detection in LLMs},
      author={Samir Abdaljalil and Hasan Kurban and Parichit Sharma and Erchin Serpedin and Rachad Atat},
      year={2025},
      eprint={2503.05980},
      archivePrefix={arXiv},
      primaryClass={cs.CL},
      url={https://arxiv.org/abs/2503.05980},
}

@misc{chen2024insidellmsinternalstates,
      title={INSIDE: LLMs' Internal States Retain the Power of Hallucination Detection},
      author={Chao Chen and Kai Liu and Ze Chen and Yi Gu and Yue Wu and Mingyuan Tao and Zhihang Fu and Jieping Ye},
      year={2024},
      eprint={2402.03744},
      archivePrefix={arXiv},
      primaryClass={cs.CL},
      url={https://arxiv.org/abs/2402.03744},
}

@misc{kim2025detectingllmhallucinationlayerwise,
      title={Detecting LLM Hallucination Through Layer-wise Information Deficiency: Analysis of Ambiguous Prompts and Unanswerable Questions},
      author={Hazel Kim and Tom A. Lamb and Adel Bibi and Philip Torr and Yarin Gal},
      year={2025},
      eprint={2412.10246},
      archivePrefix={arXiv},
      primaryClass={cs.LG},
      url={https://arxiv.org/abs/2412.10246},
}

@inproceedings{joshi-etal-2017-triviaqa,
    title = "{T}rivia{QA}: A Large Scale Distantly Supervised Challenge Dataset for Reading Comprehension",
    author = "Joshi, Mandar  and
      Choi, Eunsol  and
      Weld, Daniel  and
      Zettlemoyer, Luke",
    editor = "Barzilay, Regina  and
      Kan, Min-Yen",
    booktitle = "Proceedings of the 55th Annual Meeting of the Association for Computational Linguistics (Volume 1: Long Papers)",
    month = jul,
    year = "2017",
    address = "Vancouver, Canada",
    publisher = "Association for Computational Linguistics",
    url = "https://aclanthology.org/P17-1147/",
    doi = "10.18653/v1/P17-1147",
    pages = "1601--1611"
}

@article{NQ,
    author = {Kwiatkowski, Tom and Palomaki, Jennimaria and Redfield, Olivia and Collins, Michael and Parikh, Ankur and Alberti, Chris and Epstein, Danielle and Polosukhin, Illia and Devlin, Jacob and Lee, Kenton and Toutanova, Kristina and Jones, Llion and Kelcey, Matthew and Chang, Ming-Wei and Dai, Andrew
                        M. and Uszkoreit, Jakob and Le, Quoc and Petrov, Slav},
    title = {Natural Questions: A Benchmark for Question Answering
                    Research},
    journal = {Transactions of the Association for Computational Linguistics},
    volume = {7},
    pages = {453-466},
    year = {2019},
    month = {08},
    issn = {2307-387X},
    doi = {10.1162/tacl_a_00276},
    url = {https://doi.org/10.1162/tacl_a_00276},
    eprint = {https://direct.mit.edu/tacl/article-pdf/doi/10.1162/tacl_a_00276/1923288/tacl_a_00276.pdf},
}

@misc{rajpurkar2018knowdontknowunanswerable,
      title={Know What You Don't Know: Unanswerable Questions for SQuAD},
      author={Pranav Rajpurkar and Robin Jia and Percy Liang},
      year={2018},
      eprint={1806.03822},
      archivePrefix={arXiv},
      primaryClass={cs.CL},
      url={https://arxiv.org/abs/1806.03822},
}

@article {Krithara2022.12.14.520213,
	author = {Krithara, Anastasia and Nentidis, Anastasios and Bougiatiotis, Konstantinos and Paliouras, Georgios},
	title = {BioASQ-QA: A manually curated corpus for Biomedical Question Answering},
	elocation-id = {2022.12.14.520213},
	year = {2022},
	doi = {10.1101/2022.12.14.520213},
	publisher = {Cold Spring Harbor Laboratory},
	URL = {https://www.biorxiv.org/content/early/2022/12/16/2022.12.14.520213},
	eprint = {https://www.biorxiv.org/content/early/2022/12/16/2022.12.14.520213.full.pdf},
	journal = {bioRxiv}
}

@misc{gemmateam2024gemma2improvingopen,
      title={Gemma 2: Improving Open Language Models at a Practical Size},
      author={Gemma Team and Morgane Riviere and Shreya Pathak and Pier Giuseppe Sessa and Cassidy Hardin and Surya Bhupatiraju and Léonard Hussenot and Thomas Mesnard and Bobak Shahriari and Alexandre Ramé and Johan Ferret and Peter Liu and Pouya Tafti and Abe Friesen and Michelle Casbon and Sabela Ramos and Ravin Kumar and Charline Le Lan and Sammy Jerome and Anton Tsitsulin and Nino Vieillard and Piotr Stanczyk and Sertan Girgin and Nikola Momchev and Matt Hoffman and Shantanu Thakoor and Jean-Bastien Grill and Behnam Neyshabur and Olivier Bachem and Alanna Walton and Aliaksei Severyn and Alicia Parrish and Aliya Ahmad and Allen Hutchison and Alvin Abdagic and Amanda Carl and Amy Shen and Andy Brock and Andy Coenen and Anthony Laforge and Antonia Paterson and Ben Bastian and Bilal Piot and Bo Wu and Brandon Royal and Charlie Chen and Chintu Kumar and Chris Perry and Chris Welty and Christopher A. Choquette-Choo and Danila Sinopalnikov and David Weinberger and Dimple Vijaykumar and Dominika Rogozińska and Dustin Herbison and Elisa Bandy and Emma Wang and Eric Noland and Erica Moreira and Evan Senter and Evgenii Eltyshev and Francesco Visin and Gabriel Rasskin and Gary Wei and Glenn Cameron and Gus Martins and Hadi Hashemi and Hanna Klimczak-Plucińska and Harleen Batra and Harsh Dhand and Ivan Nardini and Jacinda Mein and Jack Zhou and James Svensson and Jeff Stanway and Jetha Chan and Jin Peng Zhou and Joana Carrasqueira and Joana Iljazi and Jocelyn Becker and Joe Fernandez and Joost van Amersfoort and Josh Gordon and Josh Lipschultz and Josh Newlan and Ju-yeong Ji and Kareem Mohamed and Kartikeya Badola and Kat Black and Katie Millican and Keelin McDonell and Kelvin Nguyen and Kiranbir Sodhia and Kish Greene and Lars Lowe Sjoesund and Lauren Usui and Laurent Sifre and Lena Heuermann and Leticia Lago and Lilly McNealus and Livio Baldini Soares and Logan Kilpatrick and Lucas Dixon and Luciano Martins and Machel Reid and Manvinder Singh and Mark Iverson and Martin Görner and Mat Velloso and Mateo Wirth and Matt Davidow and Matt Miller and Matthew Rahtz and Matthew Watson and Meg Risdal and Mehran Kazemi and Michael Moynihan and Ming Zhang and Minsuk Kahng and Minwoo Park and Mofi Rahman and Mohit Khatwani and Natalie Dao and Nenshad Bardoliwalla and Nesh Devanathan and Neta Dumai and Nilay Chauhan and Oscar Wahltinez and Pankil Botarda and Parker Barnes and Paul Barham and Paul Michel and Pengchong Jin and Petko Georgiev and Phil Culliton and Pradeep Kuppala and Ramona Comanescu and Ramona Merhej and Reena Jana and Reza Ardeshir Rokni and Rishabh Agarwal and Ryan Mullins and Samaneh Saadat and Sara Mc Carthy and Sarah Cogan and Sarah Perrin and Sébastien M. R. Arnold and Sebastian Krause and Shengyang Dai and Shruti Garg and Shruti Sheth and Sue Ronstrom and Susan Chan and Timothy Jordan and Ting Yu and Tom Eccles and Tom Hennigan and Tomas Kocisky and Tulsee Doshi and Vihan Jain and Vikas Yadav and Vilobh Meshram and Vishal Dharmadhikari and Warren Barkley and Wei Wei and Wenming Ye and Woohyun Han and Woosuk Kwon and Xiang Xu and Zhe Shen and Zhitao Gong and Zichuan Wei and Victor Cotruta and Phoebe Kirk and Anand Rao and Minh Giang and Ludovic Peran and Tris Warkentin and Eli Collins and Joelle Barral and Zoubin Ghahramani and Raia Hadsell and D. Sculley and Jeanine Banks and Anca Dragan and Slav Petrov and Oriol Vinyals and Jeff Dean and Demis Hassabis and Koray Kavukcuoglu and Clement Farabet and Elena Buchatskaya and Sebastian Borgeaud and Noah Fiedel and Armand Joulin and Kathleen Kenealy and Robert Dadashi and Alek Andreev},
      year={2024},
      eprint={2408.00118},
      archivePrefix={arXiv},
      primaryClass={cs.CL},
      url={https://arxiv.org/abs/2408.00118},
}

@misc{touvron2023llama2openfoundation,
      title={Llama 2: Open Foundation and Fine-Tuned Chat Models},
      author={Hugo Touvron and Louis Martin and Kevin Stone and Peter Albert and Amjad Almahairi and Yasmine Babaei and Nikolay Bashlykov and Soumya Batra and Prajjwal Bhargava and Shruti Bhosale and Dan Bikel and Lukas Blecher and Cristian Canton Ferrer and Moya Chen and Guillem Cucurull and David Esiobu and Jude Fernandes and Jeremy Fu and Wenyin Fu and Brian Fuller and Cynthia Gao and Vedanuj Goswami and Naman Goyal and Anthony Hartshorn and Saghar Hosseini and Rui Hou and Hakan Inan and Marcin Kardas and Viktor Kerkez and Madian Khabsa and Isabel Kloumann and Artem Korenev and Punit Singh Koura and Marie-Anne Lachaux and Thibaut Lavril and Jenya Lee and Diana Liskovich and Yinghai Lu and Yuning Mao and Xavier Martinet and Todor Mihaylov and Pushkar Mishra and Igor Molybog and Yixin Nie and Andrew Poulton and Jeremy Reizenstein and Rashi Rungta and Kalyan Saladi and Alan Schelten and Ruan Silva and Eric Michael Smith and Ranjan Subramanian and Xiaoqing Ellen Tan and Binh Tang and Ross Taylor and Adina Williams and Jian Xiang Kuan and Puxin Xu and Zheng Yan and Iliyan Zarov and Yuchen Zhang and Angela Fan and Melanie Kambadur and Sharan Narang and Aurelien Rodriguez and Robert Stojnic and Sergey Edunov and Thomas Scialom},
      year={2023},
      eprint={2307.09288},
      archivePrefix={arXiv},
      primaryClass={cs.CL},
      url={https://arxiv.org/abs/2307.09288},
}

@misc{yang2025qwen3technicalreport,
      title={Qwen3 Technical Report},
      author={An Yang and Anfeng Li and Baosong Yang and Beichen Zhang and Binyuan Hui and Bo Zheng and Bowen Yu and Chang Gao and Chengen Huang and Chenxu Lv and Chujie Zheng and Dayiheng Liu and Fan Zhou and Fei Huang and Feng Hu and Hao Ge and Haoran Wei and Huan Lin and Jialong Tang and Jian Yang and Jianhong Tu and Jianwei Zhang and Jianxin Yang and Jiaxi Yang and Jing Zhou and Jingren Zhou and Junyang Lin and Kai Dang and Keqin Bao and Kexin Yang and Le Yu and Lianghao Deng and Mei Li and Mingfeng Xue and Mingze Li and Pei Zhang and Peng Wang and Qin Zhu and Rui Men and Ruize Gao and Shixuan Liu and Shuang Luo and Tianhao Li and Tianyi Tang and Wenbiao Yin and Xingzhang Ren and Xinyu Wang and Xinyu Zhang and Xuancheng Ren and Yang Fan and Yang Su and Yichang Zhang and Yinger Zhang and Yu Wan and Yuqiong Liu and Zekun Wang and Zeyu Cui and Zhenru Zhang and Zhipeng Zhou and Zihan Qiu},
      year={2025},
      eprint={2505.09388},
      archivePrefix={arXiv},
      primaryClass={cs.CL},
      url={https://arxiv.org/abs/2505.09388},
}

@misc{tonmoy2024comprehensive,
      title={A Comprehensive Survey of Hallucination Mitigation Techniques in Large Language Models},
      author={S. M Towhidul Islam Tonmoy and S M Mehedi Zaman and Vinija Jain and Anku Rani and Vipula Rawte and Aman Chadha and Amitava Das},
      year={2024},
      eprint={2401.01313},
      archivePrefix={arXiv},
      primaryClass={cs.CL},
      url={https://arxiv.org/abs/2401.01313},
}

@article{Huang_2025,
   title={A Survey on Hallucination in Large Language Models: Principles, Taxonomy, Challenges, and Open Questions},
   volume={43},
   ISSN={1558-2868},
   url={http://dx.doi.org/10.1145/3703155},
   DOI={10.1145/3703155},
   number={2},
   journal={ACM Transactions on Information Systems},
   publisher={Association for Computing Machinery (ACM)},
   author={Huang, Lei and Yu, Weijiang and Ma, Weitao and Zhong, Weihong and Feng, Zhangyin and Wang, Haotian and Chen, Qianglong and Peng, Weihua and Feng, Xiaocheng and Qin, Bing and Liu, Ting},
   year={2025},
   month=jan, pages={1–55} }

@article{ji2023survey,
   title={Survey of Hallucination in Natural Language Generation},
   volume={55},
   ISSN={1557-7341},
   url={http://dx.doi.org/10.1145/3571730},
   DOI={10.1145/3571730},
   number={12},
   journal={ACM Computing Surveys},
   publisher={Association for Computing Machinery (ACM)},
   author={Ji, Ziwei and Lee, Nayeon and Frieske, Rita and Yu, Tiezheng and Su, Dan and Xu, Yan and Ishii, Etsuko and Bang, Ye Jin and Madotto, Andrea and Fung, Pascale},
   year={2023},
   month=mar, pages={1–38} }

@misc{zheng2023judgingllmasajudgemtbenchchatbot,
      title={Judging LLM-as-a-Judge with MT-Bench and Chatbot Arena},
      author={Lianmin Zheng and Wei-Lin Chiang and Ying Sheng and Siyuan Zhuang and Zhanghao Wu and Yonghao Zhuang and Zi Lin and Zhuohan Li and Dacheng Li and Eric P. Xing and Hao Zhang and Joseph E. Gonzalez and Ion Stoica},
      year={2023},
      eprint={2306.05685},
      archivePrefix={arXiv},
      primaryClass={cs.CL},
      url={https://arxiv.org/abs/2306.05685},
}

@article{Azamfirei2023,
      author = {Azamfirei, Razvan and Kudchadkar, Sapna and Fackler, James},
      year = {2023},
      month = {03},
      pages = {},
      title = {Large language models and the perils of their hallucinations},
      volume = {27},
      journal = {Critical Care},
      doi = {10.1186/s13054-023-04393-x}
}

@misc{lin2022teaching,
      title={Teaching Models to Express Their Uncertainty in Words}, 
      author={Stephanie Lin and Jacob Hilton and Owain Evans},
      year={2022},
      eprint={2205.14334},
      archivePrefix={arXiv},
      primaryClass={cs.CL},
      url={https://arxiv.org/abs/2205.14334}, 
}

@article{belinkov-2022-probing,
    title = "Probing Classifiers: Promises, Shortcomings, and Advances",
    author = "Belinkov, Yonatan",
    journal = "Computational Linguistics",
    volume = "48",
    number = "1",
    month = mar,
    year = "2022",
    address = "Cambridge, MA",
    publisher = "MIT Press",
    url = "https://aclanthology.org/2022.cl-1.7/",
    doi = "10.1162/coli_a_00422",
    pages = "207--219"
}

@misc{lin2024generatingconfidenceuncertaintyquantification,
      title={Generating with Confidence: Uncertainty Quantification for Black-box Large Language Models},
      author={Zhen Lin and Shubhendu Trivedi and Jimeng Sun},
      year={2024},
      eprint={2305.19187},
      archivePrefix={arXiv},
      primaryClass={cs.CL},
      url={https://arxiv.org/abs/2305.19187},
}

@inproceedings{LLM-Check,
 author = {Sriramanan, Gaurang and Bharti, Siddhant and Sadasivan, Vinu Sankar and Saha, Shoumik and Kattakinda, Priyatham and Feizi, Soheil},
 booktitle = {Advances in Neural Information Processing Systems},
 doi = {10.52202/079017-1077},
 editor = {A. Globerson and L. Mackey and D. Belgrave and A. Fan and U. Paquet and J. Tomczak and C. Zhang},
 pages = {34188--34216},
 publisher = {Curran Associates, Inc.},
 title = {LLM-Check: Investigating Detection of Hallucinations in Large Language Models},
 url = {https://proceedings.neurips.cc/paper_files/paper/2024/file/3c1e1fdf305195cd620c118aaa9717ad-Paper-Conference.pdf},
 volume = {37},
 year = {2024}
}

@misc{li2024inferencetimeinterventionelicitingtruthful,
      title={Inference-Time Intervention: Eliciting Truthful Answers from a Language Model},
      author={Kenneth Li and Oam Patel and Fernanda Viégas and Hanspeter Pfister and Martin Wattenberg},
      year={2024},
      eprint={2306.03341},
      archivePrefix={arXiv},
      primaryClass={cs.LG},
      url={https://arxiv.org/abs/2306.03341},
}

@misc{wang2024factualitylargelanguagemodels,
      title={Factuality of Large Language Models: A Survey},
      author={Yuxia Wang and Minghan Wang and Muhammad Arslan Manzoor and Fei Liu and Georgi Georgiev and Rocktim Jyoti Das and Preslav Nakov},
      year={2024},
      eprint={2402.02420},
      archivePrefix={arXiv},
      primaryClass={cs.CL},
      url={https://arxiv.org/abs/2402.02420},
}

@misc{chen2024hallucinationdetectionrobustlydiscerning,
      title={Hallucination Detection: Robustly Discerning Reliable Answers in Large Language Models},
      author={Yuyan Chen and Qiang Fu and Yichen Yuan and Zhihao Wen and Ge Fan and Dayiheng Liu and Dongmei Zhang and Zhixu Li and Yanghua Xiao},
      year={2024},
      eprint={2407.04121},
      archivePrefix={arXiv},
      primaryClass={cs.CL},
      url={https://arxiv.org/abs/2407.04121},
}

@misc{zhang2025sirenssongaiocean,
      title={Siren's Song in the AI Ocean: A Survey on Hallucination in Large Language Models},
      author={Yue Zhang and Yafu Li and Leyang Cui and Deng Cai and Lemao Liu and Tingchen Fu and Xinting Huang and Enbo Zhao and Yu Zhang and Chen Xu and Yulong Chen and Longyue Wang and Anh Tuan Luu and Wei Bi and Freda Shi and Shuming Shi},
      year={2025},
      eprint={2309.01219},
      archivePrefix={arXiv},
      primaryClass={cs.CL},
      url={https://arxiv.org/abs/2309.01219},
}

@misc{zou2025representationengineeringtopdownapproach,
      title={Representation Engineering: A Top-Down Approach to AI Transparency},
      author={Andy Zou and Long Phan and Sarah Chen and James Campbell and Phillip Guo and Richard Ren and Alexander Pan and Xuwang Yin and Mantas Mazeika and Ann-Kathrin Dombrowski and Shashwat Goel and Nathaniel Li and Michael J. Byun and Zifan Wang and Alex Mallen and Steven Basart and Sanmi Koyejo and Dawn Song and Matt Fredrikson and J. Zico Kolter and Dan Hendrycks},
      year={2025},
      eprint={2310.01405},
      archivePrefix={arXiv},
      primaryClass={cs.LG},
      url={https://arxiv.org/abs/2310.01405},
}

@misc{chern2023factoolfactualitydetectiongenerative,
      title={FacTool: Factuality Detection in Generative AI -- A Tool Augmented Framework for Multi-Task and Multi-Domain Scenarios},
      author={I-Chun Chern and Steffi Chern and Shiqi Chen and Weizhe Yuan and Kehua Feng and Chunting Zhou and Junxian He and Graham Neubig and Pengfei Liu},
      year={2023},
      eprint={2307.13528},
      archivePrefix={arXiv},
      primaryClass={cs.CL},
      url={https://arxiv.org/abs/2307.13528},
}

@inproceedings{MHAD,
  title     = {Detecting Hallucination in Large Language Models Through Deep Internal Representation Analysis},
  author    = {Zhang, Luan and Song, Dandan and Wu, Zhijing and Tian, Yuhang and Zhou, Changzhi and Xu, Jing and Yang, Ziyi and Zhang, Shuhao},
  booktitle = {Proceedings of the Thirty-Fourth International Joint Conference on
               Artificial Intelligence, {IJCAI-25}},
  publisher = {International Joint Conferences on Artificial Intelligence Organization},
  editor    = {James Kwok},
  pages     = {8357--8365},
  year      = {2025},
  month     = {8},
  note      = {Main Track},
  doi       = {10.24963/ijcai.2025/929},
  url       = {https://doi.org/10.24963/ijcai.2025/929},
}

@misc{vaswani2023attentionneed,
      title={Attention Is All You Need},
      author={Ashish Vaswani and Noam Shazeer and Niki Parmar and Jakob Uszkoreit and Llion Jones and Aidan N. Gomez and Lukasz Kaiser and Illia Polosukhin},
      year={2023},
      eprint={1706.03762},
      archivePrefix={arXiv},
      primaryClass={cs.CL},
      url={https://arxiv.org/abs/1706.03762},
}

@misc{kornblith2019similarityneuralnetworkrepresentations,
      title={Similarity of Neural Network Representations Revisited},
      author={Simon Kornblith and Mohammad Norouzi and Honglak Lee and Geoffrey Hinton},
      year={2019},
      eprint={1905.00414},
      archivePrefix={arXiv},
      primaryClass={cs.LG},
      url={https://arxiv.org/abs/1905.00414},
}

@inproceedings{wang-etal-2025-ranked,
    title = "Ranked Voting based Self-Consistency of Large Language Models",
    author = "Wang, Weiqin  and
      Wang, Yile  and
      Huang, Hui",
    editor = "Che, Wanxiang  and
      Nabende, Joyce  and
      Shutova, Ekaterina  and
      Pilehvar, Mohammad Taher",
    booktitle = "Findings of the Association for Computational Linguistics: ACL 2025",
    month = jul,
    year = "2025",
    address = "Vienna, Austria",
    publisher = "Association for Computational Linguistics",
    url = "https://aclanthology.org/2025.findings-acl.744/",
    doi = "10.18653/v1/2025.findings-acl.744",
    pages = "14410--14426",
    ISBN = "979-8-89176-256-5"
}

\appendix

\section{Implementation details}
\label{appendix:implementation_details}

\subsection{Label construction and judge protocol}

We employed different annotation strategies depending on dataset characteristics.

\paragraph{SQuAD.}
Following the official dataset evaluation protocol, we operationalize correctness using a
token-overlap F1 threshold of $50.0$; answers above the threshold are treated as correct and those
below it are treated as incorrect for label construction. This is a dataset-specific proxy rather
than a universal hallucination definition.

\paragraph{TriviaQA, NQ, and BioASQ.}
We formulate an \emph{LLM-as-a-judge} labeling protocol
\citep{zheng2023judgingllmasajudgemtbenchchatbot} using \texttt{DeepSeek-V3} as the adjudicator
for semantic equivalence. When multiple reference answers are available, we ask whether the
predicted answer means the same as any of the expected answers; when only one reference answer is
available, we ask the same question with the singular form. In both cases, the judge is required
to respond only with ``yes'' or ``no,'' following the protocol used in semantic-entropy-style
semantic equivalence evaluation.

\subsection{Probe configuration and optimization}

For any given LLM-dataset pair, we compute multivariate Fisher scores for all layers in a single
pass over training data. Unless otherwise stated, we use fixed settings across all experiments:
$K=15$ for the layer budget, $\lambda=1$ for the diversity weight, $\alpha=0.9$ for neuron
selection, MLP hidden dimensions of 1024 and 128, Adam with learning rate $\eta=0.001$, weight
decay $0.01$, and dropout probability $0.1$. We adopt $\alpha=0.9$ as a practical balance between
AUROC and feature dimensionality; this value is not claimed to be universally optimal and can be
adjusted jointly with other budget choices when different deployment constraints are preferred.

\paragraph{Compute infrastructure.}
All experiments were conducted on a single NVIDIA A100 80\,GB GPU. Hidden-state extraction and probe training for a given LLM-dataset pair typically complete within a few hours on this hardware; the exact wall-clock time varies with model size (e.g., shorter for Gemma-2-2B and longer for Qwen3-30B-A3B). The efficiency measurements reported in Section~\ref{sec:efficiency} were collected under the same hardware configuration with batch size 256.

\subsection{Multi-seed stability on Qwen models}

Table~\ref{tab:qwen_multiseed_main} summarizes the multi-seed results for Qwen3-14B and Qwen3-30B-A3B on the main in-domain benchmarks. To keep the presentation consistent with the main text, we report AUROC in percentage form here. Across both models, \textsc{QAoD} remains consistently the best-performing method, and the standard deviations are very small, typically below 0.4 AUROC points on TriviaQA, SQuAD, and NQ. In-domain gains are thus stable with respect to random initialization and training noise. SAPLMA, by contrast, shows a notably larger spread on BioASQ for Qwen3-14B, suggesting that some baselines are more sensitive to seed variation under domain shift.

Table~\ref{tab:qwen_multiseed_ood} reports the corresponding zero-shot OOD results on BioASQ. The same pattern holds: \textsc{QAoD} achieves the best mean AUROC for both Qwen models, and the variance is again extremely small. Compared with the in-domain tables, the OOD setting is slightly more variable for Qwen3-14B, but the absolute standard deviations remain modest, confirming that the cross-domain advantage of the question-orthogonal component feature is not a seed artifact.

\begin{table}[!ht]
\centering
\caption{Multi-seed in-domain AUROC on Qwen3-14B and Qwen3-30B-A3B (seeds 42 to 46, mean $\pm$ std over five runs); \textsc{QAoD} achieves the lowest variance, confirming stable training.}
\label{tab:qwen_multiseed_main}
\small
\setlength{\tabcolsep}{5pt}
\begin{tabular}{llcccc}
\toprule
\textbf{Model} & \textbf{Method} & \textbf{TriviaQA} & \textbf{SQuAD} & \textbf{NQ} & \textbf{BioASQ} \\
\midrule
\multirow{4}{*}{Qwen3-14B}
& MIND & 82.89 $\pm$ 0.36 & 72.45 $\pm$ 1.36 & 74.86 $\pm$ 1.55 & 78.38 $\pm$ 1.31 \\
& SAPLMA & 86.19 $\pm$ 0.39 & 83.68 $\pm$ 0.32 & 79.17 $\pm$ 0.61 & 78.12 $\pm$ 4.75 \\
& \textsc{QAoD} ($V_{\perp}$) & 87.25 $\pm$ 0.24 & 81.39 $\pm$ 0.37 & 79.62 $\pm$ 0.21 & 78.15 $\pm$ 0.08 \\
& \textsc{QAoD} ($Q \oplus V_{\perp}$) & 90.35 $\pm$ 0.15 & 85.51 $\pm$ 0.19 & 82.53 $\pm$ 0.18 & 85.51 $\pm$ 0.07 \\
\midrule
\multirow{4}{*}{Qwen3-30B-A3B}
& MIND & 80.38 $\pm$ 0.61 & 71.88 $\pm$ 1.00 & 75.39 $\pm$ 2.11 & 77.08 $\pm$ 0.72 \\
& SAPLMA & 86.87 $\pm$ 0.73 & 83.89 $\pm$ 0.28 & 81.15 $\pm$ 1.84 & 83.06 $\pm$ 0.87 \\
& \textsc{QAoD} ($V_{\perp}$) & 87.01 $\pm$ 0.15 & 82.54 $\pm$ 0.14 & 79.28 $\pm$ 0.11 & 79.97 $\pm$ 0.21 \\
& \textsc{QAoD} ($Q \oplus V_{\perp}$) & 91.86 $\pm$ 0.08 & 84.92 $\pm$ 0.15 & 83.90 $\pm$ 1.66 & 85.94 $\pm$ 0.06 \\
\bottomrule
\end{tabular}
\end{table}

\begin{table}[!ht]
\centering
\caption{Multi-seed zero-shot OOD AUROC on BioASQ (seeds 42 to 46, mean $\pm$ std over five runs); the OOD advantage of $\mathcal{H}_{V_\perp}$ holds consistently across seeds.}
\label{tab:qwen_multiseed_ood}
\small
\setlength{\tabcolsep}{8pt}
\begin{tabular}{lcc}
\toprule
\textbf{Method} & \textbf{Qwen3-14B} & \textbf{Qwen3-30B-A3B} \\
\cmidrule(lr){2-2}\cmidrule(lr){3-3}
& \footnotesize{BioASQ} & \footnotesize{BioASQ} \\
\midrule
MIND                              & 47.61 $\pm$ 2.57 & 64.36 $\pm$ 2.61 \\
SAPLMA                            & 51.42 $\pm$ 4.55 & 72.37 $\pm$ 4.93 \\
\textsc{QAoD} ($V_{\perp}$)       & \textbf{77.68} $\pm$ 0.36 & \textbf{76.81} $\pm$ 0.24 \\
\textsc{QAoD} ($Q \oplus V_{\perp}$) & 70.52 $\pm$ 2.15 & 73.93 $\pm$ 0.76 \\
\bottomrule
\end{tabular}
\end{table}

\end{document}